\documentclass[conference]{IEEEtran}
\usepackage{cite}
\usepackage{booktabs}
\usepackage{multirow}
\usepackage{bigstrut}
\usepackage{graphicx}
\usepackage{xcolor}
\usepackage{amsmath}
\usepackage{amssymb}
\usepackage[hyphens]{url}
\newcommand{\runin}[1]{\par\addvspace{4pt}\noindent\textit{#1}\enspace}

\title{\emph{Aligned}, \emph{Conflicting}, or \emph{Unspecified}? A Diagnostic Benchmark for Multimodal Time Series Forecasting}
\author{\IEEEauthorblockN{Xintong Wang\IEEEauthorrefmark{1},
Haihan Nan\IEEEauthorrefmark{2}, and
Ruidong Li\IEEEauthorrefmark{1}}
\IEEEauthorblockA{\IEEEauthorrefmark{1}Kanazawa University, Japan}
\IEEEauthorblockA{\IEEEauthorrefmark{2}Virginia Tech, USA}
\IEEEauthorblockA{Email: \IEEEauthorrefmark{1}wangxt@stu.kanazawa-u.ac.jp,
\IEEEauthorrefmark{2}hhnan@vt.edu,
\IEEEauthorrefmark{1}liruidong@ieee.org}}

\begin{document}
\maketitle

\begin{abstract}
Artificial Intelligence of Things (AIoT) systems increasingly combine
sensed time series with external text
such as expert reports and news, yet how reliably models use text that
agrees or conflicts with the sensed history remains unclear. We introduce the first real-world diagnostic benchmark
organized by text--history relationship: 1,762 text--window pairs across sensor telemetry (Weather), web-service
traffic (Wikipedia), and market data streams (Finance), each labeled \emph{Aligned},
\emph{Conflicting}, or \emph{Unspecified} by
whether the text's direction agrees with the historical trend, opposes
it, or is not stated. Across ten pretrained models evaluated zero-shot, we investigate text selection and combination under text–history relationships (RQ1) and error diagnosis and horizon-scaled revision control (RQ2). For RQ1, we compare two types of text on paired Weather windows: one directly forecasts temperature changes, while the other describes related weather conditions, such as winds and fronts. Under \emph{Aligned}, the first type yields the lowest error for most models. Under \emph{Conflicting}, it increases most models’ error above the no-text baseline; we therefore add only its stated magnitude and timing to the second type. Compared with either type alone, this combination lowers mean absolute scaled error (MASE) for six models and raises directional accuracy (DA) for seven. For RQ2, an error decomposition attributes the error increase to
oversized revisions, and a training-free rule that limits them outperforms forecasts both with and without text for the four
models with this trade-off, lowering MASE by 4.49\%--21.98\% and raising DA
by 2.99\%--9.71\%. Our dataset is available at
\url{https://github.com/Shelton1202/ACU-Bench}.
\end{abstract}

\begin{figure*}[!t]
\centering
\includegraphics[width=\textwidth]{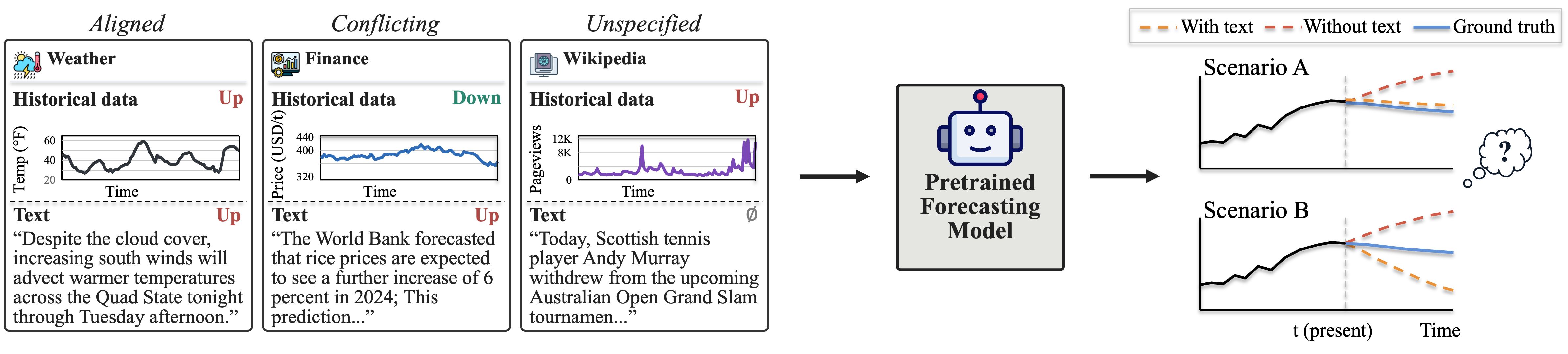}
\normalsize
\caption{Text--history relationships and possible effects of adding text. Left: \emph{Aligned} and \emph{Conflicting} indicate whether the text direction agrees with or opposes the historical trend; \emph{Unspecified} indicates that no textual direction is assigned.}
\label{fig:motivation}
\end{figure*}

\section{Introduction}

With the rapid development of Artificial Intelligence of Things (AIoT) systems, some of them rely on forecasts from
sensed time series to support
decisions. Meanwhile, Large language models (LLMs) have shown remarkable capability and are widely used in AIoT systems to forecast from serialized
values~\cite{gruver2023llmtime} and align numerical and textual
representations~\cite{jin2024timellm}, enabling forecasts conditioned on
external text such as news and reports~\cite{wu2025aurora}. Such text can
reveal changes not yet visible in the sensed history, but it can also
contradict that history. Because it typically comes from outside the
sensing pipeline, a reliable forecaster must handle disagreement between
sources rather than assume that they agree.
Consider forecasting temperature in Paducah: the recent trend was upward
when a National Weather Service report\footnote{\url{https://mesonet.agron.iastate.edu/p.php?pid=201901191150-KPAH-FXUS63-AFDPAH}} forecast the arrival of arctic air and a sharp temperature drop. The mean temperature over the next twelve hours fell, as forecast.
A forecasting model must decide what to take from this report and how
much it should change the forecast. It can predict cooling correctly and still overestimate the
temperature drop. This example raises a broader question: how does agreement or conflict
between the text and the historical trend affect forecast direction and
numerical error?

Existing evaluations approach this question by examining what textual
context contributes beyond numerical history. Time-MMD pairs real-world
series with reports across domains~\cite{liu2024timemmd}, and TimesX
broadens this evaluation to more domains and context types, including
event reports~\cite{liu2026timesx}. However, pairing text with a series
does not establish that accurate forecasting requires the text. Context
is Key addresses this issue through tasks in which textual context is
essential~\cite{williams2025context}. To identify what makes text
useful, Zhang et al.~\cite{zhang2026when} use controlled experiments to examine
whether text adds information beyond history and how explicitly it
describes future changes. CFA further examines reliability, using
controlled synthetic descriptions to test whether constrained fusion
improves robustness to incorrect or irrelevant
text~\cite{lee2026constrained}. Beyond the properties of the information
itself, What If TSF evaluates how models respond to changing scenarios:
holding history and domain background fixed, it varies the input between
factual and hypothetical scenarios and tests whether forecast revisions
satisfy expected direction and intensity
relations~\cite{jang2026whatif}.
However, none of these evaluations considers how real-world text relates
to the historical trend. A real report may agree with the trend, oppose it,
or state no direction, yet no benchmark organizes real-world evaluation
by this relationship.

To address this gap, we introduce the first real-world diagnostic
benchmark organized by text--history relationship, as illustrated in
the left panel of Fig.~\ref{fig:motivation}. We call a text \emph{directional} if
it states an increase or decrease,
and \emph{non-directional} otherwise. A directional text is
\emph{Aligned} when its direction agrees with the historical trend and
\emph{Conflicting} when it opposes the trend; a non-directional text is
\emph{Unspecified}.

We evaluate ten pretrained models zero-shot, with and without text. The
effect of text depends on the text--history relationship: adding text
lowers mean absolute scaled error (MASE) for nine models under
\emph{Aligned} but for only four under \emph{Conflicting}, even though it
still raises directional accuracy (DA) for seven. Hence adding text does
not always help, and under \emph{Conflicting} it makes forecasts
directionally more accurate yet numerically worse. These two
observations raise two questions.

\textbf{RQ1:} Which text helps under \emph{Aligned} and
\emph{Conflicting}?
Text helps fewer models under \emph{Conflicting}, raising the question
of whether a different choice or combination of text can improve
forecasting. We study this question using paired Weather segments from
the same discussion: the directional segment explicitly states a
warming or cooling to a given temperature, while the non-directional
segment describes the process behind it, such as a front's passage or a
wind shift. We first compare the two
segments with the no-text baseline under \emph{Aligned} and
\emph{Conflicting}, then test whether combining them under \emph{Conflicting} can outperform
either segment alone.

\textbf{RQ2:} Why does adding text improve DA but worsen MASE, and can
both improve?
DA checks only whether a forecast moves in the right direction, whereas
MASE also penalizes how far it moves. As in the Paducah example, a
forecast can predict cooling correctly yet overestimate the drop, so a
revision larger than needed adds error without adding DA. Likewise, DA
counts a wrong forecast as a single error however far it moves, and a
correct-to-wrong revision can cost more than a wrong-to-correct revision
saves. We first decompose the
error increase under \emph{Conflicting} by its source, then test whether
limiting large revisions can improve DA and MASE at once.

Our contributions are:
\begin{itemize}
\item \textbf{The first real-world diagnostic benchmark organized by
text--history relationship.} The benchmark has 1,762 text--window pairs
across Weather, Finance, and Wikipedia, labeled \emph{Aligned},
\emph{Conflicting}, or \emph{Unspecified}.
\item \textbf{Text selection and combination (RQ1).} Directional text helps
under \emph{Aligned} but raises MASE above the no-text baseline for seven
of ten models under \emph{Conflicting}; keeping non-directional text and
adding only the stated magnitude and timing beats either segment alone in
MASE for six of ten models and in DA for seven.
\item \textbf{Error diagnosis and revision control (RQ2).} Under \emph{Conflicting}, text corrects more forecast directions than it misdirects, but the largest revisions account for most of the increase in MASE. A training-free, horizon-scaled cap on revisions beats
forecasts both with and without text, cutting MASE by 4.49\%--21.98\% and
raising DA by 2.99\%--9.71\%.
\end{itemize}

\begin{table*}[!t]
\caption{Forecasting performance with and without text}
\label{tab:overall_window_performance}
\centering
\small
\setlength{\tabcolsep}{2pt}
\begin{tabular*}{\textwidth}{@{\extracolsep{\fill}}c*{12}{c}@{}}
\toprule
\multirow{3}{*}{Model} & \multicolumn{4}{c}{\textit{Aligned} ($n=471$)} & \multicolumn{4}{c}{\textit{Conflicting} ($n=414$)} & \multicolumn{4}{c}{\textit{Unspecified} ($n=877$)} \\
\cmidrule(lr){2-5}\cmidrule(lr){6-9}\cmidrule(lr){10-13}
 & \multicolumn{2}{c}{MASE $\downarrow$} & \multicolumn{2}{c}{DA(\%) $\uparrow$} & \multicolumn{2}{c}{MASE $\downarrow$} & \multicolumn{2}{c}{DA(\%) $\uparrow$} & \multicolumn{2}{c}{MASE $\downarrow$} & \multicolumn{2}{c}{DA(\%) $\uparrow$} \\
\cmidrule(lr){2-3}\cmidrule(lr){4-5}\cmidrule(lr){6-7}\cmidrule(lr){8-9}\cmidrule(lr){10-11}\cmidrule(lr){12-13}
 & \shortstack{w/o\\text} & \shortstack{w/\\text} & \shortstack{w/o\\text} & \shortstack{w/\\text} & \shortstack{w/o\\text} & \shortstack{w/\\text} & \shortstack{w/o\\text} & \shortstack{w/\\text} & \shortstack{w/o\\text} & \shortstack{w/\\text} & \shortstack{w/o\\text} & \shortstack{w/\\text} \\
\midrule
GPT-6.1 Sol & 0.7015 & \textbf{0.6754} & 67.73 & \textbf{71.76} & 0.6432 & \textbf{0.6366} & \textbf{61.11} & 58.94 & 0.6563 & \textbf{0.6491} & 64.20 & \textbf{65.56} \\
GPT-6 Luna & 0.7486 & \textbf{0.7350} & 56.48 & \textbf{65.82} & 0.7234 & \textbf{0.7058} & 55.56 & \textbf{57.00} & 0.7189 & \textbf{0.7108} & 53.02 & \textbf{55.76} \\
Claude Sonnet 5.5 & 0.7182 & \textbf{0.6682} & 59.66 & \textbf{73.67} & 0.7141 & \textbf{0.6924} & 56.28 & \textbf{60.87} & 0.7108 & \textbf{0.7004} & 53.82 & \textbf{56.90} \\
Claude Haiku 5.5 & 0.7792 & \textbf{0.7329} & 52.87 & \textbf{62.63} & \textbf{0.7382} & 0.7661 & \textbf{56.76} & 54.35 & \textbf{0.7430} & 0.7437 & \textbf{51.77} & 50.97 \\
Qwen3-8B & \textbf{0.9574} & 1.0848 & 43.74 & \textbf{51.59} & \textbf{0.8569} & 1.0759 & 46.14 & \textbf{54.35} & 0.9690 & \textbf{0.9402} & 44.47 & \textbf{45.27} \\
TableGPT2-7B & 0.7980 & \textbf{0.7881} & \textbf{52.44} & 51.59 & \textbf{0.7561} & 0.7906 & 47.58 & \textbf{49.76} & \textbf{0.9044} & 0.9235 & \textbf{45.27} & 44.93 \\
Aurora & 1.3631 & \textbf{1.3387} & 42.04 & \textbf{43.10} & 1.3018 & \textbf{1.2901} & 45.41 & \textbf{46.86} & \textbf{1.6532} & 1.7787 & \textbf{48.57} & 48.46 \\
TimeOmni-1-7B & 0.7976 & \textbf{0.7890} & 48.62 & \textbf{49.47} & \textbf{0.8105} & 0.8477 & 41.30 & \textbf{48.55} & 0.9350 & \textbf{0.9154} & 40.02 & \textbf{41.28} \\
DoubleCast & 0.8002 & \textbf{0.7895} & 58.39 & \textbf{58.81} & \textbf{0.7463} & 0.7514 & \textbf{56.28} & \textbf{56.28} & 0.7051 & \textbf{0.7023} & 55.64 & \textbf{56.10} \\
TimeBraid-2.5B & 0.7918 & \textbf{0.7766} & 58.39 & \textbf{60.51} & \textbf{0.7585} & 0.7741 & 54.83 & \textbf{56.52} & 0.8378 & \textbf{0.8314} & \textbf{54.96} & 54.85 \\
\midrule
Improved models & \multicolumn{2}{c}{9/10} & \multicolumn{2}{c}{9/10} & \multicolumn{2}{c}{4/10} & \multicolumn{2}{c}{7/10} & \multicolumn{2}{c}{7/10} & \multicolumn{2}{c}{6/10} \\
\bottomrule
\end{tabular*}
\normalsize

\end{table*}

\section{Benchmark Construction}
\label{sec:benchmark}
\subsection{Data Collection}
\label{sec:data_sources}
We collect time series and associated texts from three domains:
Weather, Finance, and Wikipedia. Each variable corresponds to a
temperature series from one station, a commodity-price or exchange-rate
series, or a pageview series for one Wikipedia page. Each instance
pairs the 96 observations immediately before a forecast origin with
a text to predict the next 12 observations. The dataset contains
1,762 text--window pairs across 1,396 numerical windows, with
366 Weather windows each paired with two segments.

\runin{Weather.}
We collect hourly air temperatures in degrees Fahrenheit from ASOS observations\footnote{\url{https://mesonet.agron.iastate.edu/request/download.phtml}} at five stations
and forecaster-written NWS Area Forecast Discussions\footnote{\url{https://mesonet.agron.iastate.edu/wx/afos/}}
through the IEM archive. Origins span 2004--2026.

\runin{Finance.}
We use daily and weekly series of commodity prices and exchange rates
from TimesX~\cite{liu2026timesx}, paired with its event descriptions.
Origins for the 64 variables span 2022--2026.

\runin{Wikipedia.}
We collect daily pageviews from the Web Traffic Time Series Forecasting
dataset~\cite{kaggle2017webtraffic} and the Wikimedia Pageviews API\footnote{\url{https://doc.wikimedia.org/generated-data-platform/aqs/analytics-api/reference/page-views.html}}, with
related passages from Wikipedia and Wikinews. Origins for the 445 person pages span 2015--2018.

\subsection{Historical Direction and Window Screening}
\label{sec:construction}
Given history $x_{i,1:96}$, we determine the historical direction by
comparing the means of the final two 12-step segments,
\[
\Delta_i=\overline{x}_{i,85:96}-\overline{x}_{i,73:84},
\]
whose sign defines $H_i\in\{\mathrm{up},\mathrm{down}\}$. To keep $H_i$
unambiguous, we retain only histories with pronounced changes:
$|\Delta_i|$ must be at least $5.4^{\circ}$F for Weather, and
$|\Delta_i|/\overline{x}_{i,73:84}$ at least 3.00\% for Finance commodity
prices, 0.60\% for Finance exchange rates, or 15.00\% for Wikipedia
pageviews. Thresholds are domain-specific because the sources differ in scale and
volatility; they were calibrated on the candidate pool so that each domain
retains a comparable number of windows while the required change stays well
above typical one-step fluctuations. To ensure a consistent recent trend, a linear fit to the final
24 historical observations must have a slope with the sign of $H_i$ and
$R^2\geq0.35$. 

\subsection{Text Context Construction}
\label{sec:text_context}
We extract Weather segments from discussions issued before the forecast
origin. Finance texts are TimesX event descriptions. Wikipedia passages
describe events concerning the person, with pronouns replaced by names
and dates added, and no other edits. Every text carries evidence of
publication before the forecast origin.

Two independent GPT-6 Astra annotation runs assign a textual direction
to each text, and a third run reviews both annotations and determines
the final textual direction $T_i\in\{\mathrm{up},\mathrm{down},\varnothing\}$,
indicating an increase, decrease, or no stated direction. Annotation never sees the numerical history or the future. Texts
whose textual direction cannot be resolved are excluded. The textual
direction is the literal sign of an observed or expected change stated
in the text. Weather segments refer to the forecast horizon, so their
direction is a forecast; Finance texts may refer to a period within or
outside the horizon, so their direction may describe a change already
observed. Non-directional text states no such direction; it may
describe wind conditions, for example.

Combining $T_i$ with $H_i$ gives three conditions: \emph{Aligned} when
$T_i=H_i$, \emph{Conflicting} when the two directions are opposite, and
\emph{Unspecified} when $T_i=\varnothing$. These conditions compare
signs and do not judge whether the text is correct. \emph{Aligned} comprises 201 Weather and 270 Finance pairs,
\emph{Conflicting} 165 Weather and 249 Finance pairs, and
\emph{Unspecified} 401 Weather and 476 Wikipedia pairs. Paired Weather windows have a
directional and a non-directional segment from the same discussion, with
identical history and target; 35 Weather windows have no directional
segment and retain only the non-directional one. In paired Weather
comparisons, each window's condition is determined by the directional
segment's relation to history.

\section{Experiments and Findings}

\subsection{Evaluated Models}
\label{sec:evaluated_models}
We evaluate ten pretrained models. Six are language models: the five general-purpose LLMs GPT-6.1 Sol\footnote{\url{https://developers.openai.com/api/docs/models/gpt-6.1-sol}}, GPT-6 Luna\footnote{\url{https://developers.openai.com/api/docs/models/gpt-6-luna}}, Claude Sonnet 5.5\footnote{\url{https://www.anthropic.com/claude-sonnet-5-5}}, Claude Haiku 5.5\footnote{\url{https://www.anthropic.com/claude-haiku-5-5}}, and Qwen3-8B~\cite{yang2025qwen3}, plus the table-specialized TableGPT2-7B~\cite{su2024tablegpt2}. Four are time series models: Aurora~\cite{wu2025aurora}, TimeOmni-1-7B~\cite{guan2025timeomni}, DoubleCast~\cite{zheng2026doublecast}, and TimeBraid-2.5B~\cite{wang2026timebraid}.

The models generate zero-shot forecasts using pretrained checkpoints or API
endpoints, with no training, fine-tuning, or in-context
demonstrations on our benchmark.
\begin{table*}[!t]
\caption{Forecasting performance on paired Weather windows}
\label{tab:weather_dual_text}
\centering
\small
\setlength{\tabcolsep}{0.2pt}
\begin{tabular*}{\textwidth}{@{\extracolsep{\fill}}l*{12}{c}@{}}
\toprule
\multirow{3}{*}{Model} & \multicolumn{6}{c}{\textit{Aligned} ($n=201$)} & \multicolumn{6}{c}{\textit{Conflicting} ($n=165$)} \\
\cmidrule(lr){2-7}\cmidrule(lr){8-13}
 & \multicolumn{3}{c}{MASE $\downarrow$} & \multicolumn{3}{c}{DA(\%) $\uparrow$} & \multicolumn{3}{c}{MASE $\downarrow$} & \multicolumn{3}{c}{DA(\%) $\uparrow$} \\
\cmidrule(lr){2-4}\cmidrule(lr){5-7}\cmidrule(lr){8-10}\cmidrule(lr){11-13}
 & \footnotesize\shortstack{w/o\\text} & \footnotesize\shortstack{w/\\Directional} & \footnotesize\shortstack{w/ Non-\\directional} & \footnotesize\shortstack{w/o\\text} & \footnotesize\shortstack{w/\\Directional} & \footnotesize\shortstack{w/ Non-\\directional} & \footnotesize\shortstack{w/o\\text} & \footnotesize\shortstack{w/\\Directional} & \footnotesize\shortstack{w/ Non-\\directional} & \footnotesize\shortstack{w/o\\text} & \footnotesize\shortstack{w/\\Directional} & \footnotesize\shortstack{w/ Non-\\directional} \\
\midrule
GPT-6.1 Sol & 0.3539 & \textbf{0.2658} & 0.3239 & 85.07 & \textbf{96.02} & 86.57 & \textbf{0.2904} & 0.3600 & 0.3078 & 87.27 & 85.45 & \textbf{88.48} \\
GPT-6 Luna & 0.6111 & \textbf{0.3718} & 0.5147 & 61.19 & \textbf{86.07} & 62.69 & 0.5017 & 0.5270 & \textbf{0.4332} & 76.36 & 77.58 & \textbf{78.18} \\
Claude Sonnet 5.5 & 0.6046 & \textbf{0.3502} & 0.4787 & 62.19 & \textbf{91.04} & 71.64 & 0.5565 & 0.5487 & \textbf{0.5210} & 76.97 & \textbf{80.00} & 76.97 \\
Claude Haiku 5.5 & 0.7245 & \textbf{0.4944} & 0.6312 & 56.22 & \textbf{70.65} & 54.73 & \textbf{0.4640} & 0.6093 & 0.4876 & \textbf{76.36} & 75.15 & 75.76 \\
Qwen3-8B & 0.8575 & \textbf{0.8202} & 0.8892 & 42.29 & \textbf{58.71} & 43.28 & \textbf{0.5042} & 0.7663 & 0.5341 & 48.48 & \textbf{67.27} & 50.91 \\
TableGPT2-7B & \textbf{0.6222} & 0.7233 & 0.6262 & 52.24 & 51.24 & \textbf{55.22} & \textbf{0.5087} & 0.6434 & 0.5181 & 61.21 & \textbf{67.27} & 58.79 \\
Aurora & 0.8472 & \textbf{0.8345} & 0.8349 & 44.78 & \textbf{45.77} & 45.27 & 0.8160 & \textbf{0.8005} & 0.8011 & \textbf{60.00} & \textbf{60.00} & \textbf{60.00} \\
TimeOmni-1-7B & \textbf{0.8139} & 0.8320 & 0.8927 & 43.78 & \textbf{50.75} & 41.29 & \textbf{0.6531} & 0.6881 & 0.6668 & 47.27 & \textbf{68.48} & 49.09 \\
DoubleCast & 0.4407 & \textbf{0.4064} & 0.4287 & 68.16 & \textbf{74.13} & 68.16 & \textbf{0.3919} & 0.4019 & 0.3941 & \textbf{79.39} & 78.18 & \textbf{79.39} \\
TimeBraid-2.5B & 0.4623 & \textbf{0.4235} & 0.4486 & 61.69 & \textbf{67.66} & 62.69 & 0.4308 & 0.4287 & \textbf{0.4134} & 78.18 & 78.18 & \textbf{78.79} \\
\bottomrule
\end{tabular*}
\normalsize

\end{table*}

\begin{figure*}[!t]
\centering
\includegraphics[width=\textwidth]{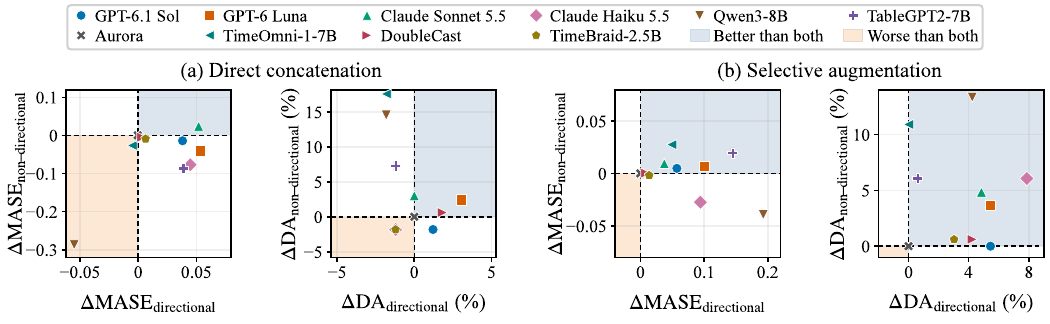}
\normalsize
\caption{Gains of (a) direct concatenation and (b) selective augmentation on 165 \emph{Conflicting} Weather windows, relative to directional text horizontally and non-directional text vertically. $\Delta\mathrm{MASE}$ is single-text minus combined-text MASE. $\Delta\mathrm{DA}$ is combined-text minus single-text DA.}
\label{fig:combination_gains}
\end{figure*}

\subsection{Evaluation Protocols}
\label{sec:evaluation_protocols}

\runin{Accuracy metrics.}
Let $\widehat y_i$ and $y_i$ be the forecast and observed future series
of window $i$, and $\ell_i=\|\widehat y_i-y_i\|_1/(12s_i)$ its scaled
error, where $s_i$ is the mean absolute seasonal difference in the
window's history. Mean absolute scaled error (MASE) sums $\ell_i$ per
variable, divides it by the same sum for the seasonal-naive forecast,
and takes the geometric mean across variables~\cite{liu2026timesx}.
Directional accuracy (DA) is the percentage of text--window pairs whose
forecast direction matches the observed one; no-text forecasts are scored
on the corresponding numerical windows. Directions compare the mean of
$\widehat y_i$ or $y_i$ with the mean of the last 12 historical values
and are classified as upward, downward, or stable using domain-specific
thresholds. Lower MASE and higher DA are better.

\runin{Forecast revisions.}
The normalized forecast revision magnitude is
$R_i=\|\widehat y_i^D-\widehat y_i^N\|_1/(12s_i)$, where
$\widehat y_i^D$ and $\widehat y_i^N$ are forecasts with directional text
and without text.

\runin{Error contributions.}
To identify which window groups drive the MASE change, we sum
their weighted changes in scaled forecast error:
\begin{equation}
C_g=\sum_{i\in g}w_{v(i)}(\ell_i^D-\ell_i^N).
\label{eq:exact_error_contribution}
\end{equation}
Here $\ell_i^D$ and $\ell_i^N$ are the scaled errors with directional
text and without text, and $v(i)$ is the window's variable. Positive weights $w_v$, one per variable, make the
contributions sum exactly to the MASE with text minus the MASE without
text. Positive $C_g$ adds error; negative $C_g$ reduces it.

\subsection{Overall Results}
\label{sec:overall_performance}

Forecasts with and without text are compared in
Table~\ref{tab:overall_window_performance}. Under \emph{Aligned}, text raises DA for nine of ten
models and lowers MASE for nine. Under \emph{Conflicting}, text still
raises DA for seven models but lowers MASE for only four. Under
\emph{Unspecified}, whose text states no direction and therefore cannot
conflict with the history, text raises DA for six models and lowers
MASE for seven. Two observations follow. First, adding text does not
always help. RQ1 therefore asks which text, and which combination of
texts, helps under each condition. Second, under \emph{Conflicting}, text makes forecasts directionally more accurate yet numerically worse;
RQ2 therefore asks why, and whether both metrics can improve.

\subsection{RQ1: Which Text Helps under \textit{Aligned} and \textit{Conflicting}?}
\label{sec:rq1_input}

\runin{Which segment helps under each condition?}
Forecasts without text, with the directional segment, and with the
non-directional segment are compared in Table~\ref{tab:weather_dual_text}
on paired Weather windows. Under
\emph{Aligned}, the directional segment yields the lowest MASE of the
three inputs for eight of ten models and the highest DA for nine. Under
\emph{Conflicting}, the same segment raises MASE above the no-text
baseline for seven models, whereas the non-directional segment stays close
to it and yields lower MASE than the directional segment for nine; DA
favors neither segment.

\runin{Why neither segment suffices under \textit{Conflicting}.}
The two segments fall short in complementary ways. The directional
segment states the direction and the expected magnitude and timing of the
target's change but not the process behind it. Because MASE scores the
whole 12-step path whereas DA checks only the mean direction, a model
that adopts the right direction can still forecast the wrong path. The
non-directional segment describes that process but does not state that
the target will change, so models have no basis for departing from the
trend. Under
\emph{Aligned}, the stated change continues the historical trend, and
models can infer the shape of the path from the history. Under
\emph{Conflicting}, the historical trend opposes the text's direction,
and the process described in the report is needed to determine the path
of the change. We therefore hypothesize that forecasting under
\emph{Conflicting} requires both the statement of the target's change and
the process that shapes its path. Since each segment carries one of the
two, we first test the simplest combination, direct concatenation.

\runin{Does direct concatenation outperform both single segments?}
As shown in Fig.~\ref{fig:combination_gains}(a), on the 165
\emph{Conflicting} windows, concatenation yields lower MASE than both
single segments for only one model and higher DA than both for only two:
supplying both segments does not realize their
complementarity. A possible reason is that the full directional segment
tends to induce oversized revisions under \emph{Conflicting}, and direct
concatenation retains this segment together with its statements of
direction. On the other hand, the expected magnitude and timing in the
directional segment may still provide useful information for the
forecast. We therefore test a selective combination: the non-directional
segment as the base, supplemented only with these details.

\runin{Can selective augmentation outperform both single segments?}
Using a fixed prompt, GPT-6 Astra keeps the non-directional segment unchanged and appends at
most one verbatim span of the directional segment stating timing,
magnitude, or a numerical constraint. It excludes direction-only
additions and spans with ambiguous scope or compatibility; it uses only
the supplied segments. With no eligible addition, the original
non-directional text is used directly.

As shown in Fig.~\ref{fig:combination_gains}(b), this input yields lower
MASE than both single segments for six of ten models and higher DA than
both for seven. This supports the hypothesis that forecasting under
\emph{Conflicting} needs both the stated change and the process behind
it, and shows that the gain appears only when the change enters as
magnitude and timing rather than as the full directional segment.

\runin{Finding 1.}
The same directional segment is the best input for most models under
\emph{Aligned} but raises error under \emph{Conflicting}, where the
non-directional segment is largely harmless. Under \emph{Conflicting},
neither segment suffices alone: keeping the non-directional segment and
adding only the magnitude and timing of the stated change lowers MASE below both
single segments for six of ten models and raises DA above both for
seven.
\par

\begin{table}[!t]
\caption{Directional-correctness transitions on \emph{Conflicting} windows}
\label{tab:bq1_transitions}
\centering\small\setlength{\tabcolsep}{2pt}
\begin{tabular*}{\columnwidth}{@{\extracolsep{\fill}}l*{6}{c}@{}}
\toprule
Model & \multicolumn{2}{c}{\shortstack{Wrong-to-\\Correct}} & \multicolumn{2}{c}{\shortstack{Correct-to-\\Wrong}} & \multicolumn{2}{c}{\shortstack{Subtotal}} \\
 & $n$ & $C_g$ & $n$ & $C_g$ & $n$ & $C_g$ \\
\midrule
Qwen3-8B & 89 & $-0.0782$ & 55 & $+0.1706$ & 144 & $+0.0923$ \\
TableGPT2-7B & 48 & $-0.0239$ & 39 & $+0.0492$ & 87 & $+0.0254$ \\
TimeOmni-1-7B & 68 & $-0.0445$ & 38 & $+0.0660$ & 106 & $+0.0215$ \\
TimeBraid-2.5B & 29 & $-0.0213$ & 22 & $+0.0218$ & 51 & $+0.0004$ \\
\bottomrule
\end{tabular*}
\par\vspace{6pt}
\begin{tabular*}{\columnwidth}{@{\extracolsep{\fill}}l*{6}{c}@{}}
\toprule
Model & \multicolumn{2}{c}{\shortstack{Correct-to-\\Correct}} & \multicolumn{2}{c}{\shortstack{Wrong-to-\\Wrong}} & \multicolumn{2}{c}{\shortstack{Subtotal}} \\
 & $n$ & $C_g$ & $n$ & $C_g$ & $n$ & $C_g$ \\
\midrule
Qwen3-8B & 136 & $+0.0350$ & 134 & $+0.0916$ & 270 & $+0.1266$ \\
TableGPT2-7B & 158 & $+0.0001$ & 169 & $+0.0090$ & 327 & $+0.0091$ \\
TimeOmni-1-7B & 133 & $-0.0030$ & 175 & $+0.0187$ & 308 & $+0.0157$ \\
TimeBraid-2.5B & 205 & $+0.0177$ & 158 & $-0.0025$ & 363 & $+0.0152$ \\
\bottomrule
\end{tabular*}
\normalsize

\end{table}

\subsection{RQ2: Why Does Adding Text Improve DA but Worsen MASE, and Can Both Improve?}
\label{sec:rq2_output}

We examine all 414 \emph{Conflicting} windows, 165 from Weather and 249
from Finance, for the four models whose DA rises but MASE worsens in
Table~\ref{tab:overall_window_performance}: Qwen3-8B, TableGPT2-7B,
TimeOmni-1-7B, and TimeBraid-2.5B.

\runin{Where does the MASE increase arise?}
As shown in Table~\ref{tab:bq1_transitions}, the windows are partitioned
into four groups by whether the forecast direction is correct before and
after adding text.
By Eq.~(\ref{eq:exact_error_contribution}), the four groups'
contributions sum to the MASE change. For all four models, more windows
turn from wrong to correct than from correct to wrong, so DA rises.

MASE rises for two reasons, one in each half of
Table~\ref{tab:bq1_transitions}: both subtotals are positive for all
four models, and together they equal the net MASE increase. First, the
windows whose correctness changes add net error even though
wrong-to-correct windows outnumber correct-to-wrong windows. The reason
is that the mean contribution per window, $C_g/n$, is 1.35 to 3.53 times larger in
magnitude for correct-to-wrong windows than for wrong-to-correct windows.
Second, the windows whose correctness is unchanged add net error that DA
cannot register: in correct-to-correct windows the revision can overshoot
the observed change, and in wrong-to-wrong windows it can move an already
wrong forecast farther from the observed values.

A revision can add no more error than the distance it moves the
forecast. This holds in all three cases above: flipping a correct
direction, overshooting in correct-to-correct windows, and moving
farther in wrong-to-wrong windows. Small revisions therefore add little
error, so we hypothesize that the added error concentrates in the
largest revisions.

\begin{table}[!t]
\caption{Contributions $C_g$ by revision quartile on \emph{Conflicting} windows}
\label{tab:revision_contribution}
\centering\small\setlength{\tabcolsep}{2pt}
\begin{tabular*}{\columnwidth}{@{\extracolsep{\fill}}lrr@{}}
\toprule
Model & Top 25\% & Rest 75\% \\
\midrule
Qwen3-8B & $+0.1971$ & $+0.0219$ \\
TableGPT2-7B & $+0.0298$ & $+0.0047$ \\
TimeOmni-1-7B & $+0.0316$ & $+0.0056$ \\
TimeBraid-2.5B & $+0.0095$ & $+0.0061$ \\
\bottomrule
\end{tabular*}
\normalsize
\end{table}


\runin{Is the net MASE increase concentrated in the largest revisions?}
We rank each model's 414 \emph{Conflicting} windows by revision magnitude
$R_i$ and compare the contributions of the largest quartile (104 windows)
with those of the remaining 310. As shown in
Table~\ref{tab:revision_contribution}, the largest quartile
accounts for 89.98\%, 86.49\%, 84.98\%, and 60.74\% of the net MASE increase for
Qwen3-8B, TableGPT2-7B, TimeOmni-1-7B, and TimeBraid-2.5B, respectively,
while holding 25.12\% of the windows. Large revisions are therefore harmful
on net under \emph{Conflicting}. This motivates limiting large revisions
while leaving small ones intact, and asking whether the directional
gains survive.

\begin{table}[!t]
\caption{Forecasting performance with revision control on \emph{Conflicting} windows}
\label{tab:revision_capping_comparison}
\centering\small\setlength{\tabcolsep}{1pt}
\begin{tabular*}{\columnwidth}{@{\extracolsep{\fill}}l*{6}{r}@{}}
\toprule
\multirow{2}{*}{Model} & \multicolumn{2}{c}{w/o text} & \multicolumn{2}{c}{w/ text} & \multicolumn{2}{c}{\shortstack{w/ revision\\control}} \\
\cmidrule(lr){2-3}\cmidrule(lr){4-5}\cmidrule(lr){6-7}
 & MASE & DA(\%) & MASE & DA(\%) & MASE & DA(\%) \\
\midrule
Qwen3-8B & 0.8569 & 46.14 & 1.0759 & 54.35 & \textbf{0.8394} & \textbf{59.18} \\
TableGPT2-7B & 0.7561 & 47.58 & 0.7906 & 49.76 & \textbf{0.7551} & \textbf{54.59} \\
TimeOmni-1-7B & 0.8105 & 41.30 & 0.8477 & 48.55 & \textbf{0.7898} & \textbf{52.66} \\
TimeBraid-2.5B & 0.7585 & 54.83 & 0.7741 & 56.52 & \textbf{0.7383} & \textbf{58.21} \\
\bottomrule
\end{tabular*}
\normalsize

\end{table}

\runin{Can limiting large revisions improve both metrics?}
The result above calls for a rule that keeps small revisions, limits
large ones, and needs neither condition labels nor tuned parameters. We
call it horizon-scaled revision control and take its scale from the
window's own history. For $L=96$ observations, the normalized mean absolute one-step change is
\begin{equation}
\nu_i=\frac{1}{(L-1)s_i}\sum_{t=1}^{L-1}|x_{i,t+1}-x_{i,t}|,
\label{eq:history_variation_reference}
\end{equation}
that is, how much the series typically moves in one step, and
$F\nu_i$ is how far it would move over the $F=12$-step horizon at that
rate. Let $\bar x_i$ be the
mean of the last $F$ historical values and
$d_i^c=|\operatorname{mean}(\widehat y_i^c)-\bar x_i|/s_i$, $c\in\{N,D\}$,
the normalized displacement of each forecast mean from it. The contraction
$\kappa_i=\max(d_i^N-d_i^D,0)$ measures how far the text pulls the
forecast mean back toward the recent history. The controlled forecast
interpolates the two forecasts:
\begin{equation}
\begin{gathered}
\widehat y_i^{\mathrm{ctrl}}=(1-a_i)\widehat y_i^N+a_i\widehat y_i^D,\\
a_i=\begin{cases}
1, & R_i\leq\nu_i,\\
\exp[-(R_i/(F\nu_i))^2]\dfrac{\nu_i}{\nu_i+\kappa_i}, & R_i>\nu_i.
\end{cases}
\end{gathered}
\label{eq:history_revision_control}
\end{equation}
In $a_i$, the first case is the gate, the exponential is the magnitude
factor, and the fraction is the contraction factor.
Small revisions ($R_i\le\nu_i$) can add only limited error, so the gate
preserves them unchanged. Larger revisions can introduce more error, but
removing them entirely would also remove any directional benefit. The
magnitude factor therefore reduces the text weight smoothly as
$R_i/(F\nu_i)$ increases. Revision magnitude alone does not
capture how much the text pulls the forecast mean back toward recent
history, and the flips and overshoots in
Table~\ref{tab:bq1_transitions} can arise when this contraction goes too far,
motivating an additional safeguard. The
contraction factor provides this safeguard by reducing the text weight
further as $\kappa_i$ increases.

Forecasts without text, with text, and with revision control are
compared in Table~\ref{tab:revision_capping_comparison}. For all four models, the
rule yields lower MASE and higher DA than forecasts both with and
without text. Relative to the unmodified text-conditioned forecasts,
MASE falls by 4.49\%--21.98\% and DA rises by 2.99\%--9.71\%. Removing either factor raises MASE for all four models.

\runin{Finding 2.}
Under \emph{Conflicting}, the MASE increase comes from oversized
text-induced revisions, whose largest quartile carries 60.74\%--89.98\% of it.
Horizon-scaled revision control, which limits revisions beyond the
history's typical change and needs neither condition labels nor tuned
parameters, yields lower MASE and higher DA than forecasts both with and
without text for all four models.

\section{Conclusion}
We introduce the first real-world diagnostic benchmark organized by
text--history relationship. Under \emph{Conflicting}, text often improves
forecast direction but increases error through oversized revisions.
Selective augmentation of non-directional text and training-free revision
control mitigate this trade-off. Future work could combine them to guide
text selection and revision across text--history relationships in AIoT
systems.

\bibliographystyle{IEEEtran}
\bibliography{references}

\end{document}